# Particle Filter SLAM for Vehicle Localization

LIU, Tianrui [1]*   XU, Changxin [2]   QIAO, Yuxin [3]   JIANG, Chufeng [4]   YU, Jiqiang [3]

[1] University of California San Diego, USA

[2] Northern Arizona University, USA

[3] Universidad Internacional Isabel I de Castilla, Spain

[4] The University of Texas at Austin, USA

*LIU, Tianrui is the corresponding author, E-mail: tianrui.liu.ml@gmail.com*

**Abstract:** Simultaneous Localization and Mapping (SLAM) presents a formidable challenge in robotics, involving the dynamic construction of a map while concurrently determining the precise location of the robotic agent within an unfamiliar environment. This intricate task is further compounded by the inherent "chicken-and-egg" dilemma, where accurate mapping relies on a dependable estimation of the robot's location, and vice versa. Moreover, the computational intensity of SLAM adds an additional layer of complexity, making it a crucial yet demanding topic in the field. In our research, we address the challenges of SLAM by adopting the Particle Filter SLAM method. Our approach leverages encoded data and fiber optic gyro (FOG) information to enable precise estimation of vehicle motion, while lidar technology contributes to environmental perception by providing detailed insights into surrounding obstacles. The integration of these data streams culminates in the establishment of a Particle Filter SLAM framework, representing a key endeavor in this paper to effectively navigate and overcome the complexities associated with simultaneous localization and mapping in robotic systems.

**Keywords:** SLAM, Computer vision, Localization, Particle Filter

## 1 Introduction

Simultaneous Localization and Mapping (SLAM) presents significant challenges in the field of robotics, requiring the construction and continuous updating of a map in an unknown environment while simultaneously tracking the agent's location. The inherent difficulty in SLAM stems from the unknown nature of both the environment and the robot's location, demanding accurate estimation. Additionally, SLAM is computationally intensive, underscoring its complexity in robotics.

In this Particle Filter SLAM framework, we specifically employ Particle Filter SLAM for efficient map construction and updates. Leveraging encoder and fiber optic gyro (FOG) data aids in precise motion estimation, while lidar technology enhances environmental perception, offering valuable insights into surrounding obstacles. Combining these data sources, our study establishes a Particle Filter SLAM framework, presenting a comprehensive approach to address the challenges of simultaneous localization and mapping in robotics.

Our endeavor demonstrates the practical impact of this research, showcasing the effectiveness of Particle Filter SLAM in solving real-world problems. By integrating diverse data streams, our approach not only advances our understanding of SLAM but also holds promise for applications in autonomous navigation, robotics, and related fields. The insights gained from this study contribute to the broader landscape of robotics research, emphasizing the practical implications and potential advancements in the realm of simultaneous localization and mapping.

## 2 Problem Formulation

In our pipeline, we adopt a common structure similar to other robotics problem. Time $t$ is discretized due to sensor observations, and the robot state $x_t$ includes position, orientation, and velocity. Control input $u_t$ governs the vehicle's motion, while observations $z_t$ consist of lidar points, texture camera data, and others. The environment state $m_t$ represents the map of space occupancy.

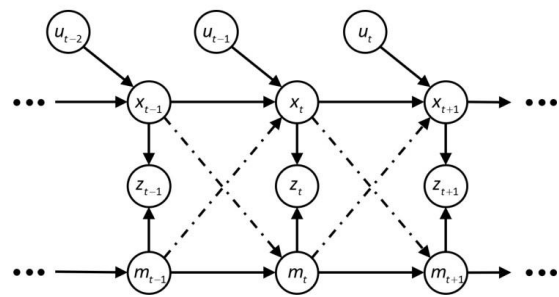

**Fig. 1. A general structure of robotic problems**

The Motion Model encapsulates the robot's movement to a new state $x_{t+1}$ after applying control input $u_t$, accounting for uncertainties denoted by $w_t$. The Observation

Model defines the observation $z_t$ based on the robot's state $x_t$ and the environment state $m_t$, considering uncertainties represented by $v_t$.

Employing the Markov assumption, we utilize the Bayes Filter, a probabilistic inference technique for state estimation in dynamic systems. The Bayes filter comprises two main steps: the Updated Step, which estimates the current state given past observations and control inputs, and the Predicted Step, which predicts the next state based on the current observations and control inputs.

Combining above we obtain the position and orientation of the vehicle at any given time.

## 3 Implementation

**Motion Model**: a nonlinear function $f$ or equivalently a probability density function pf that describes the motion of the robot to a new state

$$x_{t+1} = f(x_t, u_t, w_t)$$

**Observation Model**: a function $h$ or equivalently a probablity density function $p_h$ that describes the observation $z_t$ of the robot depending on $x_t$ and $m_t$

$$z_t = h(x_t, m_t, v_t)$$

Under Markov assumption, we can use Bayes Filter, a probabilistic inference technique for estimating the state of a dynamical system. In general, the Bayes filter comprises of two steps:

**Updated step**:

$$p_{t|t}(x_t) \coloneqq p(x_t|z_{0:t}, u_{0:t-1})$$

Update step uses scan-grid correlation to correct the robot pose.Transform the LiDAR scan to the world frame using each particle's pose hypothesis. Compute the correlation between the world-frame scan and the occupancy map. Based on the score of map correlation, we update the weights of each particle. Choosing the particle with the largest probability, we can then update the map loglikelihood.

**Resampling**: During the update, the weights of some particles will gradually become very small. In practice, we apply sample importance sampling (SIR) to the Particle Filter

**Predicted step**:

$$p_{t+1|t}(x_{t+1}) \coloneqq p(x_{t+1}|z_{0:t}, u_{0:t})$$

Using the encoder and FOG data, we can compute the instantaneous linear and angular velocities and estimate the robot trajectory via the differential drive motion model.

**Mapping**: using the lidar points, we can obtain the estimated surrounding. Worth mentioning, to obtain the correct information, we need to transform the lidar points from the lidar frame to the world frame. Using Bresengam2D algorithm, we can update the log likelihood of the occupancy map.

**Texture Map**: We use a RGB map with dimensions $3 \times length \times width$ to storing the RGB values. Using the pose of the most confident particle and having transformed the disparity map point cloud from the left camera frame to the world coordinates, we can then assign colors to the occupancy grid cells and make visualization. For each stereo image pair, the pixels whose z coordinate falls within the region of the 2D map are used to determine the map cell that this pixel is contained in. Then, we associate the RGB value of the pixel with this map cell.

## 4 Technical Approach

**Mapping**: Lidar scans in the range of [-5°, 180°], getting 286 datapoints at each timestamp. Suppose at time $t$, the lidar point positions at $\theta t$ with distance $xt$, then we can derive a point at lidar frame $(x_t cos\theta t, x_t sin\theta t)$.t the position of such point in the world frame, we have:

$$R_{world2lidar} = R_{world2vehicle} \times R_{vehicle2lidar}$$

$$P_{world2lidar} = P_{world2vehicle} + R_{world2vehicle} \times T_{vehicle2lidar}$$

Where $R, P, T$ stands for rotation, position and transition
matrix respectively.

By using:
$$P_{world} = R_{world2lidar} \times P_{lidar} + P_{world2lidar}$$

we derive the coordinates of lidar points in the world frame.

**Prediction:** The prediction part requires using differential-drive model with velocity and angular velocity from encoders and FOG. The differential-drive model states as follow:

$$\begin{aligned} v &\coloneqq v + V_{noise} \\ a_w &= a_w + a_{wnoise} \\ position &\coloneqq position + v \cdot dt \\ orientation &\coloneqq orientation + a_w \cdot dt \end{aligned}$$

Where $dt$ can be derived from the difference of two consec-utive timestamps. The $v$ and $aw$ can be derived from encoder and FOG as follow:

$$Revolution = \frac{x_t - x_{t-1}}{dt \cdot resolution}$$

$$v = revolution \times \frac{d_{left} + d_{right}}{2} \cdot 2\pi$$

$$a_w = \frac{v_{yaw}}{dt}$$

**Update**: In the update step, we need to adjust the weight and the map log-likelihood according to the Probabilistic Occupancy Grid Mapping

The observation model odds ratio is as follow:

$$g_h(z_t|m_i, x_t) = \frac{p(m_i = 1 \mid z_t, x_t)\, p(m_i = -1)}{p(m_i = -1 \mid z_t, x_t) p(m_i = 1)}$$

**Laser Correlation Model:** models a laser scan z obtained from sensor pose x in occupancy grid m based on correlation between z and $m$. We have:

$$p_h(\mathbf{z}|x, m) \propto exp\,(corr(r(z,x), m))$$

where the likelihood of a laser scan z is proportional to the correlation between the scan's world-frame projection $y = r(z, x)$ via the robot pose x and the occupancy grid m. The $y = r(z, x)$ is the grid cell indices transformed from the lidar scan.

# 4 Experiment

We use the above system to conduct experiments in a real-world scenario and obtain decent experiment results. In general, we are able obtain a satisfactory map and localization using the Particle Filter SLAM algorithm (see an example below). The inspiring experiment results show the power of particle slam algorithms.

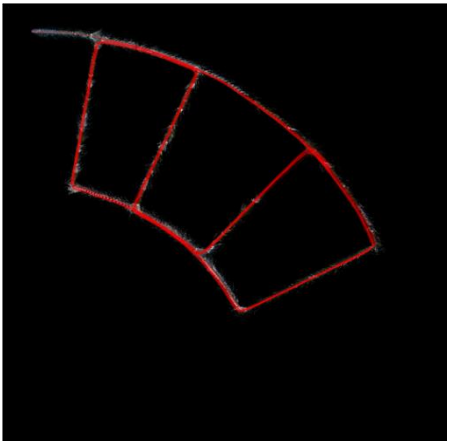

**Fig.2. Texture map trajectory**

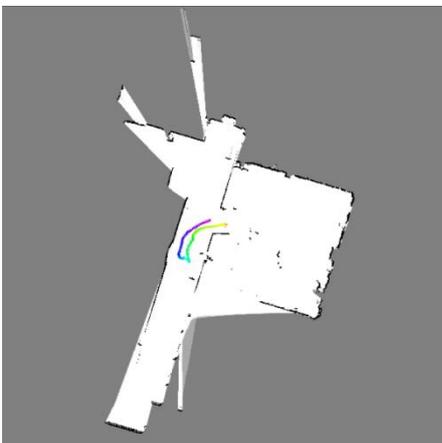

**Fig.3. Sample results of Particle Filter SLAM**

# 5 Conclusion

In summary, this study dives into an effective approach to tackle the complexities of Simultaneous Localization and Mapping (SLAM) in robotics, with a specific focus on the practical implementation of Particle Filter SLAM. By integrating encoded data, fiber optic gyro (FOG) information, and lidar technology, our Particle Filter SLAM framework offers a comprehensive solution to the challenges associated with constructing and updating maps in dynamic environments.

Navigating the uncertainties of both environmental layout and robot location, we acknowledge the computational demands inherent in SLAM. This research contributes to the field by demonstrating the practicality of Particle Filter SLAM in addressing real-world challenges. Our approach not only advances the understanding of SLAM but also holds promise for applications in autonomous navigation and robotics, showcasing its potential impact across various industries.

# Acknowledgments

The authors thank the editor and anonymous reviewers for their helpful comments and valuable suggestions.

# Funding

Not applicable.

# Institutional Review Board Statement

Not applicable.

# Informed Consent Statement

Not applicable.

# Data Availability Statement

The original contributions presented in the study are included in the article/supplementary material, further inquiries can be directed to the corresponding author.

# Conflict of Interest

The authors declare that the research was conducted in the absence of any commercial or financial relationships that could be construed as a potential conflict of interest.

# Publisher's Note

All claims expressed in this article are solely those of the authors and do not necessarily represent those of their

## Author Contributions

Not applicable.

## About the Authors

**LIU, Tianrui**

Tianrui Liu obtained his Master of Science degree in machine learning and data science from University of California San Diego. His research interests include machine learning, natural language processing, recommendation systems and robotics.

**XU, Changxin**

Affiliation: Northern Arizona University.

**QIAO, Yuxin**

Affiliation: Universidad Internacional Isabel I de Castilla, Spain.

**JIANG, Chufeng**

Affiliation: Department of Computer Science, The University of Texas at Austin.

**YU, Jiqiang**

Affiliation: Universidad Internacional Isabel I de Castilla, Spain.